\documentclass[journal=jacsat,manuscript=article]{achemso}
\usepackage[version=3]{mhchem} 
\usepackage{siunitx}
\usepackage[dvipsnames]{xcolor}
\usepackage{amssymb}
\usepackage{pifont}
\usepackage{float}
\usepackage{wrapfig}
\usepackage{hyperref}

\usepackage{amsmath}

\usepackage{tabularx,colortbl}
\usepackage{capt-of}
\usepackage{adjustbox}
\usepackage[font=normalsize,labelfont={bf},font={bf}]{caption}
\usepackage{graphicx}
\usepackage{subcaption}
\usepackage{tikz}
\usetikzlibrary{shadows}
\usepackage{caption}
\usepackage{comment}
\usepackage{array}
\usepackage{multirow}
\usepackage[labelfont=md,textfont=md]{caption}
\usepackage{algorithm}
\usepackage{algpseudocode}
\usepackage{float}
\usetikzlibrary{arrows.meta, positioning}
\usepackage{booktabs}
\usepackage{pgfplots}
\usepackage{pgfplotstable}
\pgfplotsset{compat=1.18}

\author{Srivathsan Badrinarayanan*}
\affiliation[cheme]
{Department of Chemical Engineering, Carnegie Mellon University, PA 15213, USA}

\author{Yue Su*}
\affiliation[ece]
{Department of Electrical and Computer Engineering, Carnegie Mellon University, PA 15213, USA}

\author{Janghoon Ock}
\affiliation[unl-cheme]
{Department of Chemical and Biomolecular Engineering, University of Nebraska--Lincoln, NE 68588, USA}

\author{Alan Pham}
\affiliation[ece]
{Department of Electrical and Computer Engineering, Carnegie Mellon University, PA 15213, USA}

\author{Sanya Ahuja}
\affiliation[bme]
{Department of Biomedical Engineering, Carnegie Mellon University, PA 15213, USA}

\author{Amir Barati Farimani}
\email{barati@cmu.edu}
\affiliation[meche]
{Department of Mechanical Engineering, Carnegie Mellon University, PA 15213, USA}
\alsoaffiliation[biomed]
{Department of Biomedical Engineering, Carnegie Mellon University, PA 15213, USA}
\alsoaffiliation[cheme]
{Department of Chemical Engineering, Carnegie Mellon University, PA 15213, USA}
\alsoaffiliation[mld]
{Machine Learning Department, Carnegie Mellon University, PA 15213, USA}

\title[An \textsf{achemso} demo]
{Meta-Learning for Cross-Task Generalization in Protein Mutation Property Prediction}

\begin{document}

\begin{abstract}

\noindent Protein mutations can have profound effects on biological function, making accurate prediction of property changes critical for drug discovery, protein engineering, and precision medicine. Current approaches rely on fine-tuning protein-specific transformers for individual datasets, but struggle with cross-dataset generalization due to heterogeneous experimental conditions and limited target domain data. We introduce two key innovations: (1) the first application of Model-Agnostic Meta-Learning (MAML) to protein mutation property prediction, and (2) a novel mutation encoding strategy using separator tokens to directly incorporate mutations into sequence context. We build upon transformer architectures integrating them with MAML to enable rapid adaptation to new tasks through minimal gradient steps rather than learning dataset-specific patterns. Our mutation encoding addresses the critical limitation where standard transformers treat mutation positions as unknown tokens, significantly degrading performance. Evaluation across three diverse protein mutation datasets (functional fitness, thermal stability, and solubility) demonstrates significant advantages over traditional fine-tuning. In cross-task evaluation, our meta-learning approach achieves 29\% better accuracy for functional fitness with 65\% less training time, and 94\% better accuracy for solubility with 55\% faster training.  The framework maintains consistent training efficiency regardless of dataset size, making it particularly valuable for industrial applications and early-stage protein design where experimental data is limited. This work establishes a systematic application of meta-learning to protein mutation analysis and introduces an effective mutation encoding strategy, offering transformative methodology for cross-domain generalization in protein engineering.
\end{abstract}

\section{Introduction}

Proteins are the key building blocks of life, executing a wide range of functions, including enzymatic catalysis, cellular signaling, structural support, and molecular transportation\cite{anfinsen1973principles, alberts2022molecular}. The remarkable diversity of protein functions emerges from their intricate three-dimensional structures, which are dictated by the precise order of amino acid sequences. This was a principle first established by Anfinsen\cite{anfinsen1973principles} and further explored in modern structural biology\cite{dill2007protein}. Consequently, the relationship between sequence, structure, and function underpins not only our understanding of basic biological mechanisms but also the development of therapeutic strategies and biotechnological applications\cite{huang2022advances, whisstock2003prediction}. 

Given this intricate relationship, mutations - defined as alterations to amino acid sequences - can have profound effects on protein structure and function\cite{alberts2022molecular}. While some mutations can be functionally neutral, others can cause significant disruptions, leading to cancer, genetic diseases, and neuro-degenerative disorders\cite{cooper2010proteincomplexes, pihlstrom2018genetics}. For instance, TP53 mutations alter DNA binding capability and are implicated in over 50\% of human cancers\cite{olivier2010tp53}. The challenge lies in accurately predicting these effects across diverse protein families and experimental conditions, as mutation impact depends on location, structural environment, and physicochemical properties \cite{li2014predicting, bhattacharya2017impact}. While many mutations reduce protein stability, the complex relationship between stability and functional adaptability\cite{tokuriki2009stability} adds further complexity to prediction tasks. Understanding mutation effects is essential for precision medicine, therapeutic design, and protein engineering. However, current approaches face a fundamental limitation we term the heterogeneity problem: wherein, experimental studies use varying protocols and conditions, creating unnormalized datasets where models trained on one dataset fail to generalize to others.

Traditional experimental approaches (site-directed mutagenesis, X-ray crystallography, NMR)\cite{hutchison1978mutagenesis, kendrew1958three, wuthrich1986nmr, tsai1991mechanism} provide high-resolution insights but are labor-intensive, costly, and limited in throughput\cite{sarkisyan2016local}. Early computational methods like QSAR models and FoldX\cite{guerois2002predicting, schymkowitz2005foldx} rely on handcrafted features and simplified assumptions that inadequately capture protein complexity\cite{alquraishi2019end}. Machine learning offers promising advantages: simultaneous processing of vast sequence data, identification of complex non-linear patterns, and rapid predictions once trained. Recent advances have introduced more sophisticated approaches, with deep neural networks, graph-based models, and transformers demonstrating remarkable success\cite{riesselman2018deep, gainza2020deciphering, meier2021language, jumper2021highly}.

Among these architectures, transformer-based models show particular promise for protein mutation analysis \cite{vaswani2017attention, brown2020language}. Transformers offer several key advantages over other machine learning architectures: they capture long-range amino acid dependencies, process entire sequences simultaneously, and learn context-aware representations without pre-defined structural features\cite{10.7554/eLife.82819}. Recent work has demonstrated that transformers can achieve strong property predictions from sequential representations alone, without structural inputs\cite{doi:10.1021/acs.jcim.5c01625, doi:10.1021/acscatal.3c04956, 10.1063/5.0201755, ock2024multimodal}.

Originally developed for natural language processing, transformers naturally suit protein sequences which can be represented in language-like form. Given their established ease-of-use for protein sequence modeling\cite{guntuboina2023peptidebert, doi:10.1021/acs.jcim.4c01443, doi:10.1021/acs.jpcb.4c02507, doi:10.1073/pnas.2016239118, doi:10.1126/science.ade2574, 9477085, doi:10.1021/acs.jcim.3c01706}, we adopt transformers as our base architecture.

Despite these advances, the heterogeneity problem persists as a fundamental challenge. Traditional supervised fine-tuning requires substantial target domain data and risks catastrophic forgetting of previously learned knowledge\cite{doi:10.1073/pnas.1611835114}. This limitation becomes particularly apparent when new proteins of interest have limited experimental data\cite{livesey2020using, frazer2021disease}, yet researchers need reliable predictions for mutation effects. Current approaches struggle to rapidly adapt to new protein families or experimental conditions without extensive retraining, limiting their practical utility.

\begin{figure}[ht]
    \centering
    \includegraphics[width=\textwidth]{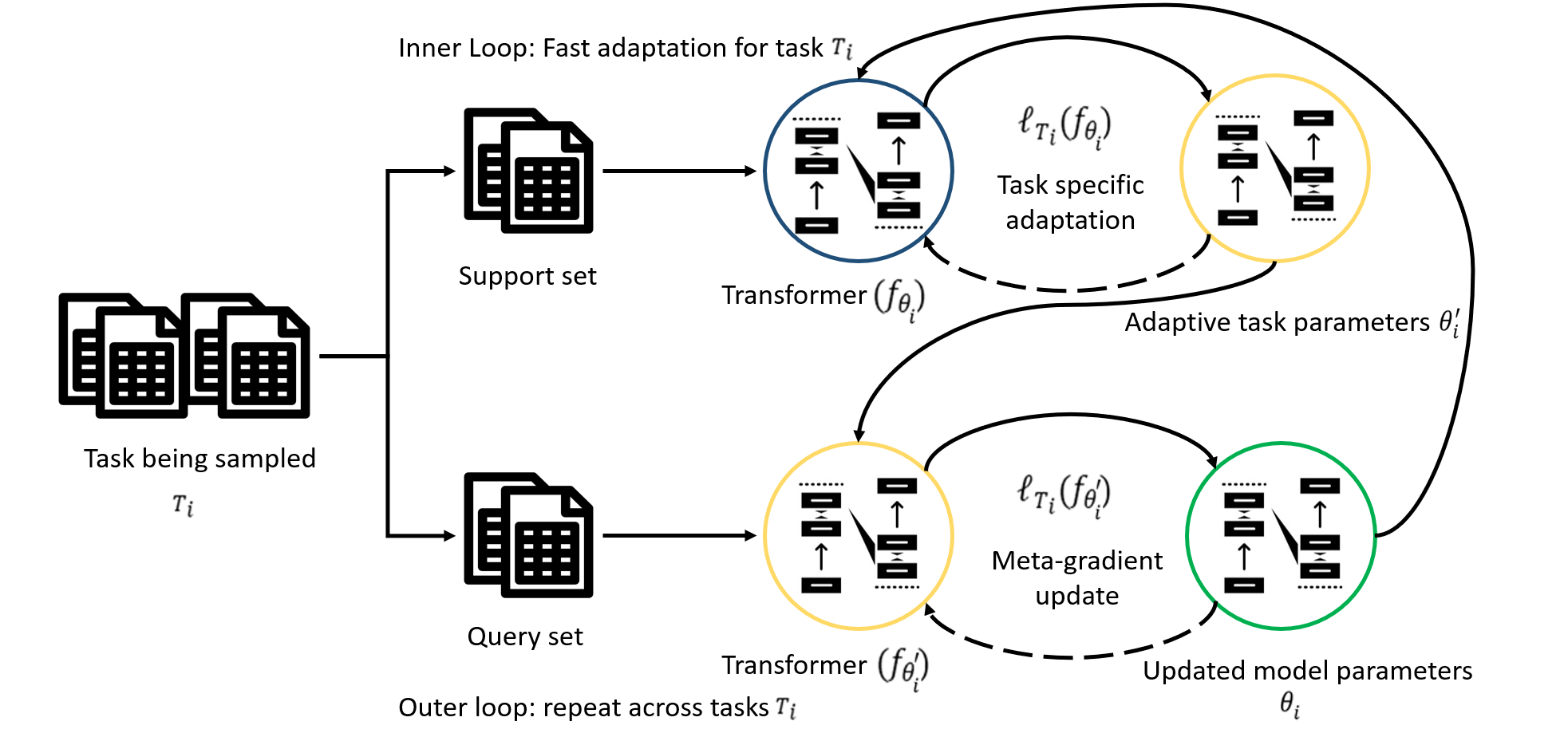}
    \caption{Overview of the MAML framework applied to protein sequence modeling. The dataset is split into support and query sets for each sampled task $\tau_i$. The support set is used in the inner loop to compute the task-specific loss and adapt the model parameters $\theta_i$ to $\theta_i'$. The query set is then used in the outer loop to compute the meta-loss and update the original model parameters $\theta_i$. The updated parameters are used to predict protein properties by feeding the data through a forward network that integrates both the protein sequences and their mutation information.}
    \label{fig:framework}
\end{figure}

To overcome this fundamental challenge, we propose two key innovations that represent the primary contributions of this work. First, we introduce the application of model-agnostic meta-learning (MAML) to protein mutation prediction\cite{pmlr-v70-finn17a}. Unlike fine-tuning that adapts all parameters to specific datasets, meta-learning trains models to be inherently adaptable, learning to adjust quickly to new tasks with minimal examples while preserving prior knowledge. This work represents a novel application of MAML principles (as shown in Figure \ref{fig:framework}) to address the cross-dataset generalization challenge in protein mutation analysis.

Second, we develop a novel mutation encoding strategy that effectively captures both sequence context and mutational changes within the transformer framework. Traditional approaches suffer from mutations being encoded as unknown tokens since positional integers lack corresponding tokens in the base transformer vocabulary. Our encoding approach enables more precise representation of mutation effects by directly incorporating mutations into sequence context, contributing to improved prediction accuracy across diverse experimental conditions.

This MAML-based computational framework addresses three critical limitations: (1) inability to generalize across heterogeneous datasets, (2) requirement for large target domain data, and (3) computational expense of conventional supervised fine-tuning. Our model leverages prior knowledge within an inner MAML training loop to adapt to predicting diverse properties such as Change in Gibbs Free Energy ($\Delta$$\Delta$G), Change in Melting Temperature ($\Delta$Tm), and Change in Solubility for any protein following mutation. 

This work advances computational protein biology through two primary contributions: bringing an application of MAML to protein mutation prediction and proposing a novel mutation encoding strategy. These innovations provide a principled approach to cross-dataset adaptation that could transform mutation effect prediction across diverse protein families. The integration of meta-learning principles with transformer frameworks provides a foundation for more generalizable mutation prediction tools, with implications for drug discovery, precision medicine, and protein engineering.

\section{Methods}

Our methodological framework integrates three key components: (1) a curated multi-source protein mutation dataset spanning diverse property types and experimental conditions, (2) a meta-learning framework based on Model-Agnostic Meta-Learning (MAML) that enables rapid adaptation to new protein property prediction tasks, and (3) an Enhanced Encoding strategy that addresses fundamental limitations in representing mutations within transformer frameworks.

\subsection{Multi-Source Dataset Integration}

To address the heterogeneity challenge outlined above and enable a robust meta-learning strategy, we constructed a unified protein mutation dataset by integrating three complementary repositories: ProteinGym Raw Substitutions Deep Mutational Scanning (DMS)\cite{NEURIPS2023_cac723e5}, FireProtDB\cite{stourac2021fireprotdb}, and SoluProtMutDB\cite{velecky2022soluprotmutdb}. These sources were strategically selected to capture diverse mutation types and experimental readouts: ProteinGym provides functional fitness scores derived from high-throughput mutational scanning assays, FireProtDB specializes in experimental protein stability measurements, and SoluProtMutDB focuses on solubility changes. Together, they span multiple mutation-induced effects, including fitness, stability, and solubility changes.

Our integrated dataset encompasses four distinct protein property types. The Functional Fitness (FF) dataset consists of 21,274 single-point mutation entries from ProteinGym, representing assay-specific functional effects relative to wild-type. For thermodynamic stability, we compiled 26,041 entries measuring Change in Gibbs Free Energy ($\Delta\Delta G$, ddG) from FireProtDB and ProteinGym, describing mutation-induced changes in thermodynamic stability. The Change in Melting Temperature ($\Delta T_m$, dTm) dataset contains 2,740 entries from FireProtDB reporting thermal stability changes. Finally, the Change in Solubility ($\Delta S$, dS) dataset comprises 16,067 multi-point mutation entries from SoluProtMutDB describing solubility changes.

\begin{figure}[ht]
    \centering
    \includegraphics[width=\textwidth]{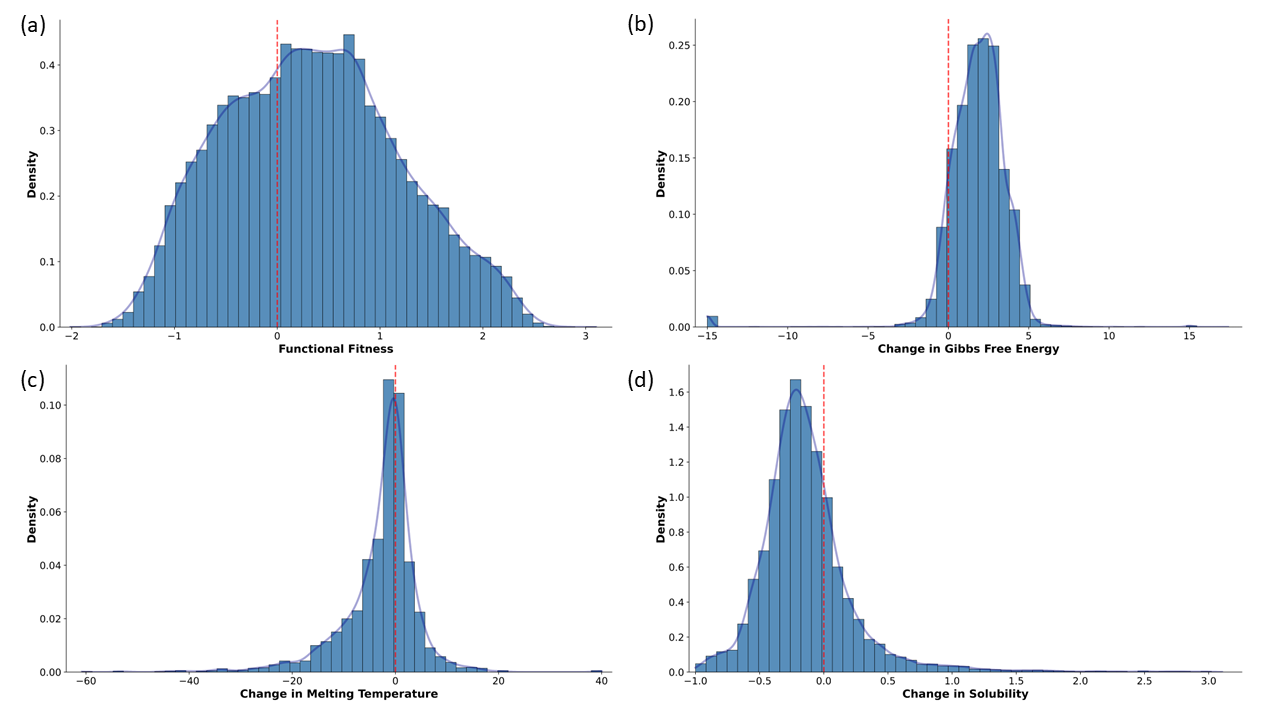}
    \caption{Distributions of protein property types in the integrated dataset: (a) Functional Fitness (FF) (b) Change in Gibbs Free Energy ($\Delta\Delta G$), (c) Change in Melting Temperature ($\Delta T_m$), and (d) Change in Solubility ($\Delta S$). Histograms are truncated at the 95th percentile for clarity; extreme outliers are excluded from visualization but retained in all analyses.}
\label{fig:dataset_distributions}
\end{figure}

After quality control and deduplication, our final dataset contained 66,122 curated mutation entries distributed across these four property categories. To illustrate the variation across datasets, we provide the raw data distributions in Figure~\ref{fig:dataset_distributions}, which highlight the strong skew toward destabilizing mutations in stability and solubility tasks, in contrast to the more balanced distribution in functional fitness assays. This is reflective of what is observed in a generalized experiment\cite{faure2015universal}. 

The measured properties span different physical scales and units: Functional Fitness values are dimensionless scores representing relative fitness normalized to wild-type, Change in Gibbs Free Energy is measured in kcal/mol, Change in Melting Temperature is expressed in degrees Celsius (\textdegree C), and Change in Solubility uses dimensionless solubility scores on a log scale. This heterogeneity in measurement scales and units, along with the varied value distributions underscores the importance of using a normalized evaluation metric, enabling fair comparison across datasets with fundamentally different physical properties. To this end, we employ the Normalized Mean Squared Error (NMSE) in this study, which is calculated as:
\begin{equation}
    NMSE = MSE / Var(y_{true})
\end{equation}

where MSE is the mean squared error and $Var(y_{true})$ is the variance of the true target values. This normalization enables fair comparison across datasets with different scales and units, ensuring that performance metrics reflect actual predictive capability rather than scale differences.

Our quality control pipeline involved sequential steps: (1) validating that amino acids are among the standard 20 and that mutation indices match corresponding residues in original sequences, and (2) performing standard normalization of the target property within each data source while preserving relative relationships within experimental datasets. This normalization strategy is critical for meta-learning success, as it maintains the integrity of experimental measurements within each source while enabling knowledge transfer across different experimental protocols and measurement scales. We addressed data redundancy by removing duplicates with identical mutation specifications. 

Each dataset entry follows a consistent structure containing the wild-type amino acid sequence, mutation specification, and corresponding target value. The specific encoding strategy for representing mutations within the transformer framework is detailed in the section titled Enhanced Mutation Encoding Strategy. 

\subsection{Meta-Learning Framework Design}

\subsubsection{Theoretical Foundation}

To address the fundamental challenge of cross-dataset generalization in protein mutation prediction, we integrate Model-Agnostic Meta-Learning (MAML)\cite{pmlr-v70-finn17a} with transformer-based architectures. Unlike traditional supervised fine-tuning approaches that adapt all parameters to individual datasets, MAML learns parameters that enable rapid adaptation to new protein property prediction tasks through minimal gradient steps by optimizing for quick task-specific parameter updates. This paradigm shift is particularly valuable for protein mutation analysis, where experimental datasets are inherently heterogeneous due to different base sequences, measurement conditions, and property types.

The meta-learning framework operates through bi-level optimization, separating learning into inner-loop task adaptation and outer-loop meta-parameter optimization. For each task $\mathcal{T}_i$ representing different experimental contexts (functional fitness, thermal stability, solubility), we formalize the adaptation process by first calculating each sampled task loss as follows:

\begin{equation}
\mathcal{L}_{\mathcal{T}_i}(f_{\theta}) = \frac{1}{|\mathcal{T}_i^{support}|}\sum_{(x^{(j)}, y^{(j)}) \sim \mathcal{T}_i^{support}} \left\| f_{\theta}(x^{(j)}) - y^{(j)} \right\|_2^2 \label{eq:1}
\end{equation}

Then we proceed with the task-specific adaptation through gradient descent on the support set:
\begin{equation}
\theta_i' = \theta - \alpha \nabla_{\theta} \mathcal{L}_{\mathcal{T}_i}(f_\theta)
\label{eq:2}
\end{equation}

The meta-objective then optimizes current task parameters $\theta$ based on adapted performance across query sets:
\begin{equation}
\theta \leftarrow \theta - \beta \nabla_\theta \sum_{\mathcal{T}_i} \mathcal{L}_{\mathcal{T}_i}(f_{\theta_i'})
\label{eq:3}
\end{equation}

where $\alpha$ and $\beta$ represent inner-loop and meta-learning rates respectively. This framework enables the model to learn generalizable representations that capture fundamental sequence-mutation-property relationships rather than dataset-specific patterns.

\subsubsection{Implementation Framework}

We utilize ProtBERT\cite{protBERT} (built over BERT\cite{BERT}) as our transformer backbone, which was then domain-adapted through supervised fine-tuning on our compiled Change in Gibbs Free Energy (ddG) dataset. This domain adaptation provides a foundation of protein property understanding that both the conventional supervised fine-tuning and our meta-learning framework builds upon. Rather than introducing architectural innovations to the transformer itself, our contributions lie in the systematic application of MAML to protein mutation prediction and our mutation encoding strategy.

Our meta-learning training protocol follows a hierarchical data splitting strategy. Each task dataset is initially split into training and testing sets using an 80-20 ratio. During meta-training, we randomly sample 16 datapoints per epoch from the training dataset, which are then divided equally into support and query sets (8 examples each). The support set is used in the inner loop for task-specific adaptation, while the query set evaluates the adapted model's performance for meta-parameter updates. This episodic training structure enables the model to learn how to rapidly adapt across diverse protein mutation prediction tasks.

Our approach fundamentally transforms protein mutation analysis from learning individual prediction tasks to learning how to learn across diverse experimental contexts. When presented with new protein families or measurement conditions, the model leverages learned meta-knowledge to achieve strong performance with minimal adaptation data, addressing the critical challenge of data scarcity in experimental protein research. Complete architectural specifications, hyperparameter configurations, and detailed implementation procedures are provided in the Supporting Information.

\subsection{Enhanced Mutation Encoding Strategy}

Our approach leverages the natural language-like properties of protein sequences for transformer processing. Proteins, like text, are sequential information where order matters critically, making them naturally compatible with transformer architectures. Each amino acid can be mapped to a unique token, allowing transformers to process protein ``sentences'' and understand how changing one ``word'' (amino acid) affects the meaning of the entire ``sentence'' (protein function). For example, a protein sequence such as \texttt{MKTAYIAKQRQISFV} can be processed directly, with mutations represented as sequence modifications (e.g., an $R \rightarrow A$ substitution at position 10).

However, a critical challenge involves designing effective input representation for protein mutations. Traditional concatenation approaches format inputs as:
\begin{equation}
\text{\texttt{[CLS] original sequence [SEP] mutation} {}}
\end{equation}

We label this as the Standard Encoding approach in this study. This type of representation presents significant challenges since mutations consist of positional integers that lack corresponding tokens in the base transformer vocabulary. This causes mutation information to be encoded as \texttt{[UNK]} (unknown) tokens, significantly degrading model performance and losing critical positional information about the mutation. While models using Standard Encoding can still generate predictions, as the transformer processes the intact wild-type sequence and learns to associate \texttt{[UNK]} tokens with sequence modifications through positional embeddings, this degraded representation fundamentally limits performance. The regression head partially compensates by extracting features from available sequence representations, but cannot fully recover the lost mutation context. 

Our Enhanced Encoding approach addresses this limitation by preserving complete mutation information within the sequence context. We propose an improved input representation that directly incorporates mutations into the sequence context. Rather than expanding the vocabulary to include integers (for mutation positions), we apply mutations directly to the original sequence at the indicated position and use \texttt{[SEP]} tokens to delineate modifications. Our enhanced format, which we call the Enhanced Encoding approach, becomes:
\begin{equation}
\text{\texttt{[CLS] sequence\_before [SEP] original\_AA [SEP] mutated\_AA [SEP] sequence\_after}}
\end{equation}

For example, given protein sequence:
\begin{center}
\texttt{SSGGSSILDRAVIEHNLLSASKLYNNITFEELGALLEIPAAKAEIIASQMITEGRMNGFIDQIDGIVHFETR}
\end{center}
and mutation \texttt{R10A} (arginine to alanine at position 10), the tokenized input becomes:
\begin{equation}
\text{\texttt{[CLS] SSGGSSILD [SEP] R [SEP] A [SEP] AVIEHN...}}
\end{equation}

\begin{figure}[ht]
    \centering
    \includegraphics[width=\textwidth]{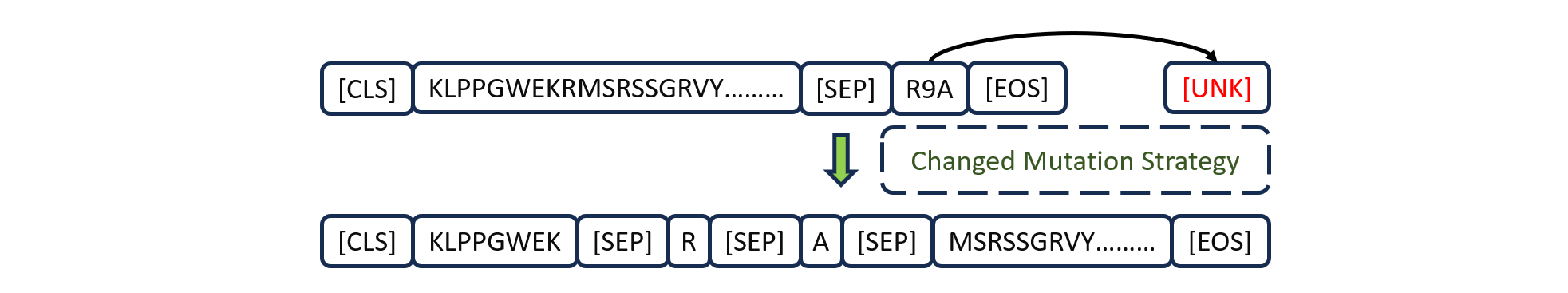}
    \caption{Comparison of input representations for encoding protein mutations in transformer models. The top row shows the Standard Encoding, where the mutation is appended as a separate string (e.g., \texttt{R9A}). This format causes issues due to the mutation position being treated as an unknown token (\texttt{[UNK]}), resulting in loss of critical positional and contextual information. The bottom row illustrates our proposed Enhanced Encoding, where the mutation is integrated directly into the sequence using \texttt{[SEP]} tokens to delineate the wild-type residue, the mutated residue, and surrounding sequence context. This approach retains full token information and improves mutation interpretability by the model. It is important to note here that even though ProtBERT tokenizes per amino acid, it is not shown in the figure as such for better clarity.}
    \label{fig:encoding}
\end{figure}

This representation clearly delineates mutation context and enables the transformer to better understand the local sequence environment. Figure \ref{fig:encoding} illustrates our mutation representation strategy.

Each dataset entry follows a consistent structure containing the wild type amino acid sequence, mutation specification, and corresponding target value indicating the property change. For our meta-learning framework, each data file was randomly split into training and testing subsets using an 80--20 split. We employ Normalized Mean Squared Error (NMSE) as the loss function to quantify differences between predicted and actual protein property changes, providing a scale-invariant measure of prediction accuracy across all target properties with different units and ranges.

\section{Results and Discussion}

We evaluated the effectiveness of our meta-learning approach through comprehensive comparison with supervised fine-tuning across three distinct protein mutation prediction tasks. Our experimental design follows two protocols: supervised fine-tuning trains models on each target dataset's training split and evaluates on the corresponding test split, while meta-learning follows a cross-task generalization protocol where models train on all other available tasks (excluding the target) and evaluate on the held-out target task's test split. This ensures meta-learning models never see target task examples during training, providing a stringent test of cross-domain generalization. For instance, when evaluating Functional Fitness (FF) prediction, the meta-learning model is trained exclusively on the complete Change in Melting Temperature (dTm) and Change in Solubility (dS) datasets, then tested on the Functional Fitness test set. Both approaches build upon ProtBERT, which was domain-adapted on the Change in Gibbs Free Energy (ddG) dataset to establish a baseline predictive model.

Table \ref{tab:cross_task_mean_var} presents our primary results comparing two meta-learning variants against traditional supervised fine-tuning, averaged across three independent trials. Meta-learning with Enhanced Encoding incorporates our novel mutation encoding strategy, while Meta-learning with Standard Encoding represents the standard MAML implementation without explicit mutation encoding. The results reveal two key findings: meta-learning consistently outperforms supervised fine-tuning across multiple tasks, and within meta-learning approaches, our mutation encoding strategy demonstrates superior performance.

\begin{table}[!ht]
\centering
\resizebox{\textwidth}{!}{%
\small
\begin{tabular}{|l|l|c|l|c|c|}
\hline
\textbf{Task} & \textbf{Method} & \multicolumn{3}{c|}{\textbf{Training}} & \textbf{NMSE} \\
\cline{3-5}
& & \textbf{Time (seconds)} & \textbf{Dataset} & \textbf{Size} & \\
\hline
\multirow{3}{*}{\shortstack[l]{Functional\\Fitness\\(FF)}} 
& Meta-learning (Enhanced) & $1497.0 \pm 18.3$ & dTm + dS & 18,807 & $\mathbf{0.73 \pm 0.00}$ \\
& Meta-learning (Standard) & $1484.3 \pm 27.6$ & dTm + dS & 18,807 & $0.75 \pm 0.00$ \\
& Supervised Fine-tuning & $4338.3 \pm 19.2$ & FF training set & 17,019 & $1.03 \pm 0.00$ \\
\hline
\multirow{3}{*}{dS} 
& Meta-learning (Enhanced) & $1489.3 \pm 3.8$ & FF + dTm & 24,014 & $\mathbf{0.06 \pm 0.00}$ \\
& Meta-learning (Standard) & $1472.7 \pm 16.5$ & FF + dTm & 24,014 & $0.09 \pm 0.00$ \\
& Supervised Fine-tuning & $3341.0 \pm 71.7$ & dS training set & 12,853 & $0.93 \pm 0.00$ \\
\hline
\multirow{3}{*}{dTm} 
& Meta-learning (Enhanced) & $1490.0 \pm 5.1$ & FF + dS & 37,341 & $1.88 \pm 0.00$ \\
& Meta-learning (Standard) & $1476.3 \pm 16.4$ & FF + dS & 37,341 & $1.91 \pm 0.00$ \\
& Supervised Fine-tuning & $\mathbf{561.7 \pm 17.6}$ & dTm training set & 2,192 & $\mathbf{0.78 \pm 0.01}$ \\
\hline
\end{tabular}
}
\caption{\textbf{Cross-task meta-learning vs. fine-tuning on three protein-related tasks (mean $\pm$ variance across three trials).} Tasks represented are: Functional Fitness (FF), Change in Melting Temperature (dTm), and Change in Solubility (dS). Meta-learning with Enhanced Encoding uses our novel mutation encoding strategy and trains on datasets excluding the target task. Meta-learning with Standard Encoding uses the standard approach without our proposed encoding. Supervised Fine-tuning trains only on the target task's training split. Bold values indicate best performance within each task for each metric. All metrics reported as Normalized Mean Squared Error (NMSE).}
\label{tab:cross_task_mean_var}
\end{table}

We observe that meta-learning demonstrates superior efficiency and accuracy compared to supervised fine-tuning. For Functional Fitness prediction, meta-learning with Enhanced Encoding achieves remarkable efficiency gains, reducing training time by 65\% (from $4338.3\pm19.2$ to $1497.0\pm18.3$ seconds) while simultaneously improving prediction accuracy by 29\% (NMSE of $0.73\pm0.00$ vs $1.03\pm0.00$). The Change in Solubility prediction task reveals even more dramatic improvements, with Meta-Learning with Enhanced Encoding reducing training time by 55\% while achieving a 94\% improvement in prediction performance (NMSE of $0.06\pm0.00$ vs $0.93\pm0.00$). These results are particularly noteworthy because meta-learning achieves superior performance despite being trained on different datasets than the target task, demonstrating effective knowledge transfer across protein property domains.

Our Enhanced Encoding strategy consistently enhances meta-learning performance across all three tasks on basis of NMSE: Functional Fitness ($0.73\pm0.00$ vs $0.75\pm0.00$), Change in Melting Temperature ($1.88\pm0.00$ vs $1.91\pm0.00$), and Change in Solubility ($0.06\pm0.00$ vs $0.09\pm0.00$). While these improvements may appear modest in absolute terms, they represent consistent gains that validate the importance of explicit mutation representation within the sequence context. The Enhanced Encoding strategy addresses a fundamental limitation of standard transformer architectures when processing protein mutations by avoiding reliance on unknown tokens and better capturing local sequence environments.

The Change in Melting Temperature (dTm) task provides important insights into meta-learning limitations. Fine-tuning achieves lower NMSE ($0.78\pm0.01$) compared to meta-learning approaches ($1.88\pm0.00$ - $1.91\pm0.00$) on this task. However, several factors explain this result: the dTm dataset is substantially smaller than other tasks (2192 vs 16000+ data points for other tasks!). Fine-tuning benefits from training directly on task-specific data, and the limited size may not provide sufficient diversity for effective cross-task learning. Notably, meta-learning maintains consistent training times across all tasks (approximately 1476-1497 seconds), demonstrating superior scalability compared to fine-tuning, whose training time varies dramatically with dataset size ($561.7\pm17.6$ seconds for Change in Melting Temperature vs $4338.3\pm19.2$ seconds for Functional Fitness). This shows the potential for meta-learning in large-scale applications.

To further investigate the potential of meta-learning when leveraging all available training data, we conducted an additional experiment where all datasets are pooled together during meta-training. Table \ref{tab:pooled_mean_var} presents results from this comprehensive evaluation, averaged across three independent trials, where meta-learning is trained on all available training splits (including the target task's training data) before evaluation on the respective test sets. This experimental design differs fundamentally from our cross-task evaluation by allowing meta-learning access to target task training data, enabling direct comparison of learning paradigms under equivalent data conditions.

\begin{table}[h]
\centering
\begin{tabular}{ |l|l|c| }
 \hline
 \textbf{Testing Dataset} & \textbf{Approach} & \textbf{NMSE} \\
 \hline
 \multirow{3}{*}{\shortstack[l]{Functional Fitness\\(FF)}} 
 & Meta-learning (pooled) & $\mathbf{0.55 \pm 0.02}$\\ 
 & Fine-tuning & $1.03 \pm 0.00$ \\
 & Fine-tuning (pooled) & $1.04 \pm 0.00$ \\
 \hline
 \multirow{3}{*}{\shortstack[l]{Change in Solubility\\(dS)}} 
 & Meta-learning (pooled) & $\mathbf{0.09 \pm 0.01}$ \\ 
 & Fine-tuning & $0.93 \pm 0.00$\\
 & Fine-tuning (pooled) & $0.94 \pm 0.00$\\
 \hline
 \multirow{3}{*}{\shortstack[l]{Change in Melting\\Temperature (dTm)}} 
 & Meta-learning (pooled) & $\mathbf{0.66 \pm 0.00}$\\ 
 & Fine-tuning & $0.80 \pm 0.01$\\ 
 & Fine-tuning (pooled) & $0.86 \pm 0.00$ \\ 
 \hline
\end{tabular}
\caption{\textbf{Comprehensive evaluation: Meta-learning vs. fine-tuning with complete dataset access (mean $\pm$ variance across three trials).} Meta-learning (pooled) trains on all available training splits from all three datasets using the MAML framework, while Fine-tuning trains only on the target task's training split. Fine-tuning (pooled) combines all three training datasets for conventional supervised fine-tuning then evaluates on each test set separately. All metrics reported as Normalized Mean Squared Error (NMSE). Training time for meta-learning (pooled) is 2,202 seconds.}
\label{tab:pooled_mean_var}
\end{table}

The pooled training experiment reveals several critical insights about the fundamental differences between meta-learning and supervised fine-tuning approaches. Most notably, meta-learning demonstrates exceptional ability to extract meaningful patterns from diverse protein mutation data, while fine-tuning struggles significantly with heterogeneous datasets. Meta-learning achieves substantial gains across all tasks: Functional Fitness ($0.55\pm0.02$ vs $1.03\pm0.00$), Change in Solubility ($0.09\pm0.01$ vs $0.93\pm0.00$), and Change in Melting Temperature ($0.66\pm0.00$ vs $0.80\pm0.01$). The pooled fine-tuning approach performs particularly poorly, suggesting that simply combining diverse datasets without appropriate learning mechanisms fails to improve protein property prediction. This limitation likely stems from fine-tuning's tendency to overfit to dominant patterns in combined datasets while failing to capture task-specific nuances.

In contrast, meta-learning's episodic training framework enables the model to learn how to rapidly adapt to different protein property prediction contexts, rather than simply memorizing task-specific patterns. This performance gap highlights a fundamental advantage of the meta-learning approach across diverse protein mutation prediction tasks.

The dTm results in the pooled experiment are particularly noteworthy, as they address the limitation observed in our cross-task evaluation. When the dTm dataset is included in meta-training (rather than excluded), meta-learning achieves excellent performance (NMSE of $0.66\pm0.00$), significantly outperforming both fine-tuning approaches ($0.80\pm0.01$ and $0.86\pm0.00$). This demonstrates that meta-learning's apparent weakness on dTm in the cross-task setting was primarily due to insufficient representation of similar tasks in the training distribution, rather than a fundamental limitation of the approach.

The combination of our cross-task and pooled training results establishes meta-learning as a superior approach for protein mutation property prediction across multiple dimensions. The methodology offers exceptional training efficiency and scalability, with consistent training times regardless of the target task or dataset size. This represents a significant practical advantage over fine-tuning, whose computational requirements scale dramatically with dataset size. This efficiency becomes increasingly valuable as protein databases continue expanding exponentially, enabling rapid deployment of prediction models for novel protein families or experimental contexts.

The approach also demonstrates robust cross-domain generalization capabilities that are essential for real-world applications. The ability to achieve superior performance on completely unseen tasks (as demonstrated in our cross-task evaluation) suggests that meta-learning captures fundamental sequence-mutation-property relationships rather than dataset-specific artifacts. This generalization capability is particularly valuable in early-stage protein design, where experimental data for novel targets may be severely limited. The integration of meta-learning with emerging agent-based frameworks\cite{zeng2025llmguidedchemicalprocessoptimization, ock2025largelanguagemodelagent, chandrasekhar2025automatingmdsimulationsproteins} could further enable autonomous protein engineering workflows, combining rapid adaptation with automated experimental design and iterative refinement.

\section{Conclusion}

This study introduces a meta-learning framework that advances protein mutation property prediction through two complementary innovations. Our primary contribution: the first systematic application of Model-Agnostic Meta-Learning (MAML) to protein mutation analysis—provides a principled framework for cross-dataset generalization that fundamentally transforms how models learn from heterogeneous experimental data. Our secondary contribution: a novel Enhanced Encoding strategy, addresses critical limitations in representing mutations within transformer architectures, amplifying the meta-learning framework's effectiveness. Together, building upon transformer architectures for protein sequence modeling, these innovations demonstrate enhanced training efficiency, robust cross-task generalization, and improved prediction accuracy compared to traditional supervised fine-tuning approaches.

Our experimental evaluation across three distinct protein mutation prediction tasks demonstrates the efficacy of meta-learning over supervised fine-tuning approaches. In cross-task evaluation scenarios, where models must generalize to completely unseen protein property prediction tasks, meta-learning with Enhanced Encoding achieves remarkable improvements: 29\% better accuracy for Functional Fitness prediction while reducing training time by 65\%, and 94\% better accuracy for solubility prediction with 55\% faster training. The pooled training experiments reveal substantial performance improvements across all tasks (47-90\%), while fine-tuning approaches show minimal gains. Notably, meta-learning resolves the Change in Melting Temperature task limitation observed in cross-task evaluation, demonstrating effective leverage of heterogeneous protein datasets.

Our Enhanced Encoding strategy represents a critical methodological contribution that consistently improves performance across all evaluated tasks. By explicitly representing mutations within sequence context rather than relying on unknown tokens, this approach addresses fundamental limitations of standard transformer architectures in protein analysis and enables better understanding of local sequence environments and mutation effects. The consistent improvements validate this design choice and demonstrate broad applicability to other transformer-based protein analysis approaches.

The practical implications are significant: the combination of improved accuracy (29-94\% in cross-task scenarios), reduced training time (55-65\% faster), and consistent computational efficiency makes meta-learning valuable for industrial applications and early-stage protein design where experimental data is limited.

In conclusion, our evaluation establishes meta-learning as a superior methodology for protein mutation property prediction, offering significant advantages in efficiency, generalization, and practical applicability. These advances provide a foundation for developing more sophisticated prediction tools that rapidly adapt to emerging techniques and novel protein families.

\section{Data and Software Availability}
The necessary code and data can be found here:
\url{https://github.com/kuku0202/maml.git}

\begin{acknowledgement}
We gratefully acknowledge the developers and maintainers of ProteinGym, FireProtDB, and SoluProtMutDB for making their curated datasets publicly available, which were essential for this work. We thank the creators of ProtBERT for developing and sharing their pre-trained protein language model. This work builds upon the foundational contributions of Finn et al. in developing the Model-Agnostic Meta-Learning (MAML) framework. Additionally, we express our gratitude to our colleagues for their valuable feedback and suggestions, which have greatly improved the quality of this work.

\end{acknowledgement}

\bibliography{references}

\end{document}


\newpage

\section{Mathematical Framework and Theoretical Foundation}
\subsection{Complete Theoretical Framework of MAML}
Model-Agnostic Meta-Learning (MAML) addresses the fundamental challenge of learning from limited data by sampling the tasks, adapting the tasks parameters and optimizing model parameters based on rapid adaptation. For protein mutation prediction, MAML proves particularly valuable given the heterogeneous nature of experimental datasets and the frequent need to predict properties differences for mutated protein sequences with minimal experimental data. The approach fundamentally differs from traditional machine learning by learning how to learn new tasks efficiently. Below are some important steps:

\textbf{Batch Task Processing:}
We use batch size equal to 4 to prevent memory overflow while processing task batches simultaneously.

\textbf{Multi-Step Inner Loop:} The adaptation phase performs multiple gradient steps (default: 5) rather than the single step often described in theoretical treatments:
\begin{equation}
\theta_i^{(k+1)} = \theta_i^{(k)} - \alpha \nabla_{\theta_i^{(k)}} \mathcal{L}_{\mathcal{T}_i}^{\text{support}}(f_{\theta_i^{(k)}})
\end{equation}

\textbf{Simplified Meta-Gradient:} Due to memory constraints with large transformer models, the meta-gradient computation employs a practical approximation. The implementation maintains full computational graphs through multiple adaptation steps, but defaults to first-order approximation for memory efficiency. The meta-gradient computation uses the standard MAML approach with gradient clipping for training stability.

The complete training procedure follows on the Algorithm 1: For each meta-training epoch, each batch tasks data is split evenly into support sets and query sets.  Each task undergoes adaptation on its support set through multiple gradient steps, producing task-adapted parameters. Then the meta-update is computed based on query set performance, with gradient clipping applied to ensure training stability.

\begin{algorithm}[H]
\caption{Model-Agnostic Meta-Learning(MAML)}
\begin{algorithmic}[1]
\Require $p(\mathcal{T})$: distribution over tasks
\Require $\alpha$, $\beta$: step size hyperparameters
\State Randomly initialize model parameters $\theta$
\While{not done}
    \State Sample a batch of tasks $\mathcal{T}_i \sim p(\mathcal{T})$
    \ForAll{tasks $\mathcal{T}_i$}
        \State Split data into support set $\mathcal{D}^{\text{train}}_i$ and query set $\mathcal{D}^{\text{test}}_i$
        \State Compute task loss on support set:
        \[
        \mathcal{L}_{\mathcal{T}_i}(f_\theta) = \frac{1}{|\mathcal{D}^{\text{train}}_i|} \sum_{(x, y) \in \mathcal{D}^{\text{train}}_i} \ell(f_\theta(x), y)
        \]
        \State Compute adapted parameters using gradient descent:
        \[
        \theta_i' = \theta - \alpha \nabla_\theta \mathcal{L}_{\mathcal{T}_i}(f_\theta)
        \]
        \State Update meta-parameters $\theta$ using query sets:
    \[
    \theta \leftarrow \theta - \beta \nabla_\theta \sum_{\mathcal{T}_i} \mathcal{L}_{\mathcal{T}_i}(f_{\theta_i'})
    \]
    \EndFor
\EndWhile
\label{algor:1}
\end{algorithmic}
\end{algorithm}

\section{Enhanced Mutation Encoding Strategy}

Our enhanced mutation encoding strategy addresses vocabulary limitations through direct sequence modification. The implementation creates a structured representation that maintains compatibility with standard protein transformer vocabularies while preserving complete mutation information.

The encoding process operates by identifying the mutation site within the protein sequence and creating explicit segments around the alteration. For a mutation R10A in the sequence \texttt{SSGGSSILDRAVIEHNLLSAS}, the transformation produces:

\texttt{[CLS] SSGGSSILD [SEP] R [SEP] A [SEP] AVIEHNLLSAS}

This segmentation approach provides several advantages. All sequence components remain within the standard amino acid vocabulary, eliminating unknown tokens while preserving complete mutation context. The explicit separation enables the transformer's attention mechanism to directly model relationships between the original residue, replacement residue, and surrounding sequence context. It prevents information loss during tokenization. Additionally, the sequences exceeding the limit undergo truncation on mutation in traditional method while mutation part is at the last position. Our mutation encoding strategy ensures mutation-relevant information remains available to the model even for long protein sequences. For proteins containing multiple mutations, the encoding system extends through additional segmentation: [CLS] seq1 [SEP] orig1 [SEP]
mut1 [SEP] seq2 [SEP] orig2 [SEP] mut2 [SEP] final seq. This approach maintains
explicit representation of each alteration while preserving the overall sequence context necessary for understanding cumulative effects. 

\section{Model Architecture and Training Configuration}
\subsection{Architecture Specifications}
The complete architecture integrates ProtBERT as the foundational transformer backbone with task-specific prediction components designed for protein property regression. ProtBERT provides pre-trained representations learned from large-scale protein sequence data, capturing fundamental patterns in amino acid co-occurrence and sequence-structure relationships.

\begin{table}[h]
\centering
\caption{Neural Network Architecture Specifications}
\label{tab:architecture}
\begin{tabular}{@{}p{4cm}p{5cm}p{6cm}@{}}
\toprule
Component & Specification \\
\midrule
Base Model & ProtBERT (110M parameters) \\
First Hidden Layer & 256 units, ReLU activation \\
Second Hidden Layer & 128 units, ReLU activation \\
Regularization & Dropout ($p=0.1$) after each layer\\
Output Layer & 1 unit, linear activation \\
\bottomrule
\end{tabular}
\end{table}

\subsection{Meta-Learning Training Protocol}
The MAML training configuration balances adaptation effectiveness with computational efficiency through carefully tuned hyperparameters.

\begin{table}[h]
\centering
\caption{MAML Training Configuration}
\label{tab:maml_config}
\begin{tabular}{@{}p{6cm}p{4cm}@{}}
\toprule
Parameter & Value \\
\midrule
Inner-loop learning rate ($\alpha$) & 0.01 \\
Meta-learning rate ($\beta$) & 0.001 \\
Support set size per task & 8 examples \\
Query set size per task & 8 examples \\
Meta-batch size & 4 \\
Meta-training episodes & 50 epochs \\
Optimizer & Adam (both loops) \\
Adaptation steps per task & 5 \\
\bottomrule
\end{tabular}
\end{table}

The inner-loop learning rate of 0.01 enables meaningful adaptation to new tasks by using the gradient of loss in support sets. The meta-learning rate of 0.001 controls updates to the meta-parameters based on query set performance across tasks, ensuring stable convergence of the meta-optimization process.

\section{Dataset Details and Preprocessing}
\subsection{Data Sources and Collection}
The training and evaluation datasets were compiled from multiple experimental sources, each providing measurements of specific protein properties under controlled conditions. Table~\ref{tab:datasets} summarizes the key characteristics of each dataset used in our study.

\begin{table}[h]
\centering
\caption{Dataset    Summary and Characteristics}
\label{tab:datasets}
\begin{tabular}{@{}p{7cm}p{2cm}p{5cm}p{3cm}p{3cm}@{}}
\toprule
Task Type & Size & Source  \\
\midrule
Binding Affinity & 21274 & ProteinGym  \\
Change in Gibbs Free Energy(ddG) & 26041 & ProteinGym, FireProt  \\
Change in Melting Temperature(dTm) & 2740 & FireProt  \\
Solubility Change  & 16067 & SoluProtMut  \\
\bottomrule
\end{tabular}
\end{table}

Each dataset underwent rigorous quality control to ensure data integrity and consistency. Sequences with non-standard amino acids, unclear mutation annotations, or missing experimental values were excluded from the analysis.

\subsection{Data Processing Pipeline}
\begin{table}[h]
\centering
\caption{Data Processing and Quality Control Parameters}
\label{tab:data_processing}
\begin{tabular}{@{}p{6cm}p{6cm}@{}}
\toprule
Parameter & Specification \\
\midrule
Maximum sequence length & 1024\\
Target normalization & StandardScalar   \\
Amino acid validation & Standard 20 proteinogenic amino acids only \\
Quality control & Removal of invalid sequences \\
Mutation validation & Verification of position-mutation correspondence \\
Train/test split & 80/20 stratified by property value \\
Cross-validation & 5-fold for hyperparameter tuning \\
\bottomrule
\end{tabular}
\end{table}

Target normalization through StandardScalar within each source ensures consistency across diverse property types and measurement scales. It prevents properties with larger numerical ranges from dominating the loss function while maintaining relative relationships within each experimental dataset.